\pdfoutput=1
\documentclass[11pt]{article}

\usepackage{acl}
\usepackage{times}
\usepackage{latexsym}
\usepackage{enumitem}

\usepackage[T1]{fontenc}
\usepackage[utf8]{inputenc}
\usepackage{xparse}
\NewDocumentCommand{\avi}
{ mO{} }{\textcolor{red}{\textsuperscript{\textit{Avi}}\textsf{\textbf{\small[#1]}}}}

\newcommand{\tydi}{{\sc TyDi QA }}

\usepackage{microtype}
\usepackage{booktabs}
\usepackage{graphicx}

\definecolor{darkgreen}{rgb}{0.0, 0.5, 0.0}

\title{Do Answers to Boolean Questions Need Explanations? Yes}

 \author{Sara Rosenthal, Mihaela Bornea, \\ \textbf{Avirup Sil}, \textbf{Radu Florian} \and \textbf{Scott McCarley}  \\
        IBM Research AI \\ \{sjrosenthal, mabornea, avi, raduf, jsmc\} @us.ibm.com}

\begin{document}
\maketitle
\begin{abstract}

Existing datasets that contain boolean questions, such as BoolQ and \tydi, provide the user with a YES/NO response to the question. However, a one word response is not sufficient for an \textit{explainable} system. We promote explainability by releasing a new set of annotations marking the evidence in existing \tydi and BoolQ datasets. We show that our annotations can be used to train a model that extracts improved evidence spans compared to models that rely on existing resources. We confirm our findings with a user study which shows that our extracted evidence spans enhance the user experience. We also provide further insight into the challenges of answering boolean questions, such as passages containing conflicting YES and NO answers, and varying degrees of relevance of the predicted evidence.

%The BoolQ dataset is considered a benchmark dataset for classifying boolean questions, answers to which is typically either \textit{yes} or \textit{no}. However, this task restriction only deals with  binary classification (YES/NO) without considering the classifier's prediction confidence and also ignores unanswerable questions. We introduce the concept of ``WEAK" confidence in addition to YES/NO labels to provide finer-grained annotations that include evidence on the sentence-level for the BoolQ and TyDi datasets further allowing multiple answers to a question in a single passage. We show the value our annotations provide to users and how they can be incorporated into an MRC system to provide improvements on popular datasets on answering boolean questions.

% The BoolQ dataset is considered a benchmark \avi{removed: the premier} dataset for classifying boolean questions, answers to which is typically either \textit{yes} or \textit{no}. However, it focuses only on YES/NO classification and ignores unanswerable questions and the confidence of the YES/NO label being correct. We introduce the concept of ``WEAK" confidence in addition to YES/NO labels to provide finer-grained annotations that include evidence on the sentence level for the BoolQ and TyDi datasets further allowing multiple answers to a question in a single passage. We show the value our annotations provide to users and how they can be incorporated into an MRC system to provide improvements on popular datasets.

\end{abstract}

\section{Introduction}

Answering questions that prompt a ``YES/NO'' response is a necessary capability of a machine reading comprehension (MRC) \cite{naturalQuestions,clark-etal-2019-boolq,clark-etal-2020-tydi} system, but
a one word response is not sufficient for an \textit{explainable system}. For example, the question ``\textit{Can you buy Cadburys creme eggs all year round?}" represents a complex information need. A simple response of ``NO'' leaves an information-hungry user of an MRC system wondering when Cadbury creme eggs can be bought. 
% Figure~\ref{fig:bool-explainable-examples} shows additional boolean questions requiring answers with explanation in the form of supporting evidence.

\begin{figure}[t]
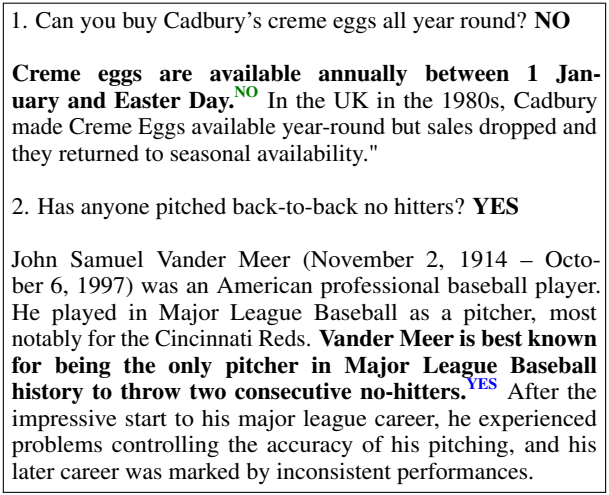

    \centering
    \small
\framebox{%
  \begin{minipage}{\columnwidth}

1. Can you buy Cadbury's creme eggs all year round?  \textbf{NO} \\
\\
\textbf{Creme eggs are available annually between 1 January and Easter Day.}\textcolor{darkgreen}{\textbf{\textsuperscript{NO}}} In the UK in the 1980s, Cadbury made Creme Eggs available year-round but sales dropped and they returned to seasonal availability." \\
\\ 
2. Has anyone pitched back-to-back no hitters? \textbf{YES} \\
\\
John Samuel Vander Meer (November 2, 1914 -- October 6, 1997) was an American professional baseball player. He played in Major League Baseball as a pitcher, most notably for the Cincinnati Reds. \textbf{Vander Meer is best known for being the only pitcher in Major League Baseball history to throw two consecutive no-hitters.}\textcolor{blue}{\textbf{\textsuperscript{YES}}} After the impressive start to his major league career, he experienced problems controlling the accuracy of his pitching, and his later career was marked by inconsistent performances.

  \end{minipage}}    
  \caption{Examples of boolean questions and their answers that desire an explanation as show in the provided passage. Supporting evidence for each answer is highlighted in bold in the passage. \textcolor{blue}{\textbf{\textsuperscript{YES}}}, and \textcolor{darkgreen}{\textbf{\textsuperscript{NO}}}}
    \label{fig:bool-explainable-examples}
\end{figure}

A successful user experience for boolean questions involves three parts: 1) The user must receive a simple YES/NO answer. This has been the focus of boolean datasets and leaderboards. 2) The user must receive supporting evidence for the answer. Most boolean questions datasets provide the full paragraph. 3) The supporting evidence must provide a positive user experience. A good evidence span quickly guides the user to a concise explanation of the answer. (See examples in Figure~\ref{fig:bool-explainable-examples}.) While the full paragraph provides some explainability, the user is forced to filter out irrelevant information. In addition, providing the degree of relevance of the evidence (highly relevant evidence will be information-rich while weakly relevant evidence may require additional inference from the user) and indicating multiple correct evidence spans can also aid in providing a positive user experience. Figure~\ref{fig:bool-examples}.1 shows an example with both strong and weak degrees of relevance. Figure~\ref{fig:bool-examples}.2 shows an example with conflicting answers.

% Explainability is a component which has been neglected in many recent boolean question answering datasets \avi{which recent datasets? Pls cite them. How is your work different than ones that do some explaination: e.g. AmbigQA or "show your work" paper or "break it down" paper. How are you different than the Checklist paper if at all.}. 
%Although a question may prompt a yes/no answer, a simple
%yes/no answer is necessary but not sufficient. Explainability #is needed to understand the YES/NO response by the MRC system. See example ....
% Many boolean questions arise from more complex information needs: the single word yes/no, (which literally answers the question), is, in practice, the start of a more verbose answer.
% see examples.
% In human conversation, a yes/no response, without elaboration, may be regarded as curt or rude.
% For an information seeking user of a machine reading comprehension (MRC) system, a one-word answer may 
% be unsatisfying, incomplete, or even lead to distrust in the system.

% Most recent question answering datasets \cite{} have focused on achieving state of the art results for questions that have short factoid answers that are a few words. However, answers to questions can come in many forms such as lists, tables, and boolean answers. In this paper, we focus on questions that are boolean such as \textit{``Have the Colorado Rockies won a title?"}. Boolean questions are often ignored by reading comprehension systems.boolean questions are important;

Existing MRC datasets do not support all aspects of this user experience. Natural Questions \cite{naturalQuestions}, \tydi \cite{clark-etal-2020-tydi}, and MS Marco \cite{DBLP:conf/nips/NguyenRSGTMD16} contain from 1-10\% boolean questions. Boolean questions have also been explored individually in the BoolQ dataset \cite{clark-etal-2019-boolq}. 
%This dataset is only designed for YES/NO answer prediction and is addressed as a label classification problem.
%In MS Marco, a naturally curated MRC dataset, over 7\% of questions from Bing query logs are YES/NO questions.
However, these datasets make several design assumptions during annotation that simplify the task thereby reducing human understanding of boolean questions: 1) Concise evidence is not provided to explain the YES/NO response, 2) The response does not incorporate the relative relevance of the answer, and 3) There is only one correct answer. Because of these limitations, MRC systems~\cite{Zhang2021PoolingformerLD,wang2020answer,Wang2020ClusterFormerCS} largely ignore boolean questions. %, even though with appropriate training they can provide evidence spans and YES/NO answers. 
% \avi{I don't understand this sentence. All MRC tasks are text classification.}. 

%  Evidence vs Factoid answer. Highlighting. Define. \avi{??}

\begin{figure}[t]
    \centering
    \small
\framebox{%
  \begin{minipage}{\columnwidth}
% \textbf{1. QUESTION:} Did the Equal Pay Act pass? \textcolor{blue}{\textbf{YES}} \\

% \textbf{PASSAGE:} The Equal Pay Act of 1963 is a United States labor law amending the Fair Labor Standards Act, aimed at abolishing wage disparity based on sex (see Gender pay gap). \textcolor{blue}{\textbf{It was signed into law on June 10, 1963}}, by John F. Kennedy as part of his New Frontier Program. 
% \textcolor{blue}{\textbf{In passing the bill}}, Congress stated that sex discrimination: depresses wages and living standards for employees necessary for their health and efficiency; prevents the maximum utilization of the available labor resources; tends to cause labor disputes, thereby burdening, affecting, and obstructing commerce; burdens commerce and the free flow of goods in commerce; and constitutes an unfair method of competition .

1. Did Avicenna write any books? \textbf{YES} \\

Avicenna 
%(/ˌævƗ'sƐnƏ/; also Ibn Sīnā or Abu Ali Sina; Persian: ابن سىنا ; c.980- June 1037) 
was a Persian polymath who is regarded as one of the most significant physicians, astronomers, thinkers and writers of the Islamic Golden Age. He has been described as the father of early modern medicine. \textbf{Of the 450 works he is known to have written}\textsuperscript{\textcolor{cyan}{WEAK\_YES}}, around 240 have survived, including 150 on philosophy and 40 on medicine.\textbf{ His most famous works are The Book of Healing, a philosophical and scientific encyclopedia, and The Canon of Medicine, a medical encyclopedia}\textsuperscript{\textcolor{blue}{YES}} which became a standard medical text at many medieval universities[15] and remained in use as late as 1650...
 \\
 
2. Was there a year 0? \textbf{YES}/\textbf{NO}\\

 \textbf{Year zero does not exist in the anno Domini system}\textsuperscript{\textcolor{darkgreen}{NO}} usually used to number years in the Gregorian calendar and in its predecessor, the Julian calendar. In this system, the year 1 BC is followed by AD 1. \textbf{However, there is a year zero in astronomical year numbering and in ISO 8601:2004 as well as in all Buddhist and Hindu calendars}\textsuperscript{\textcolor{blue}{YES}}. Historical, astronomical and ISO year numbering systems Historians. The Anno Domini era was introduced in 525 by Scythian monk Dionysius Exiguus, who used it to identify the years on his Easter table... 
  \end{minipage}}
  \caption{Examples illustrating the design assumptions that have been ignored in popular datasets containing boolean questions. All the evidence for each answer is highlighted in the passage. The superscripts indicate the evidence response type: \textcolor{blue}{\textbf{\textsuperscript{YES}}}, \textcolor{cyan}{\textbf{\textsuperscript{WEAK YES}}}, and \textcolor{darkgreen}{\textbf{\textsuperscript{NO}}.}}
    \label{fig:bool-examples}
\end{figure}

In this paper, we explore improving user understanding through more explainable answers to boolean questions in popular datasets. Our contributions are: 

% present comprehensive annotations on boolean questions that include one or more evidence spans and their corresponding fine-grained labels for a <Question, Passage> pair in existing TyDi and BoolQ datasets. Our contributions in the paper are as follows:

\begin{enumerate}
    \item \textit{Comprehensive Annotations:} We present comprehensive 
    % \avi{Please tone this down: we produce only 500 annotations.}
    % \sara{Its 3000+ annotations. Comprehensive also refers to the complexity of our anntoations} 
    annotations on over 3,000 boolean questions from existing \tydi and BoolQ datasets that add multiple supporting evidence spans to the YES/NO response as well as relevance of the supporting span. The new annotations highlight the lack of complexity found in these existing datasets. We will be releasing our annotations for research purposes. 
    %\footnote{Link will be included upon acceptance.}.
    % that cover conflicting answers in a single passage, 5-way fine-grained annotations on the sentence level consisting of a span as ``evidence" and a corresponding label. 
    \item \textit{Evidence Extraction:} Our main focus is on extracting evidence spans of text to be presented to the user to improve confidence in the system; We show that incorporating supporting evidence for boolean questions in MRC models achieves a 6 point F1 improvement in span extraction. To the best of our knowledge, we are the first to explore  evidence span extraction for boolean questions. In contrast, prior work only uses evidence during training to improve yes/no classification. %   We distinguish this from using the evidence spans by the system during training, but are not intended to be presented to the user.% To add later: and the YES/NO response for that evidence. 
    % for multiple languages and datasets: TyDi, NQ, and MS Marco.
    \item  \textit{User Experience:} Our user study shows the positive impact that supporting evidence extracted by the MRC model provides to users enhancing their experience. Receiving concise evidence in addition to the YES/NO answer and paragraph increases the confidence of the user in the answer.
\end{enumerate}

The rest of the paper is outlined as follows: We first discuss related work in Section~\ref{sec:relatedwork}. In Section~\ref{sec:data} we talk about the data followed by our annotation process in Section~\ref{sec:annotation}. We discuss our experiments and user study in Section \ref{sec:experiments} and \ref{sec:userstudy}.
Sections \ref{sec:unanswerable}, \ref{sec:yn}, and \ref{sec:weak} analyze unanswerable questions, conflicting answers, and weakly relevant answers. 
\section{Related Work}
\label{sec:relatedwork}

In addition to BoolQ \cite{clark-etal-2019-boolq}, datasets focusing on boolean questions exist in other languages such as Russian \cite{glushkova2020danetqa} and Chinese \cite{ReCoWang2020}. We only focus on English evaluation. StrategyQA \cite{geva2021strategyqa}
is a boolean multi-hop question answering dataset requiring inference over non-boolean questions. Numerous works have explored various approaches for improving results on BoolQ but have not explored the role of supporting evidence \cite{dzendzik-etal-2020-q,micheli2021structural, ross2021tailor,9237654}. 

Some papers use the spans of text internally to classify the YES/NO response of the question, but do not learn to predict evidence spans or analyze how useful evidence is to an end user.
For example, the ReCo dataset \cite{ReCoWang2020} includes evidence annotations for opinionated boolean questions in Chinese. %The dataset is collected from query logs and the focus is on opinionated boolean questions. 
Niu et al. \shortcite{niu2020selftraining} use a self-training approach to create synthetic evidence for boolean questions using the gold YES/NO answer and sentences in the passage.

%We leave evaluating our approach on the ReCo dataset as future work.
%We only look at English evaluation for now. ReCo is only Chinese. It is collected in a different style - from query logs (not Wikipedia) we leave it as future work. Do not evaluate how they do on the evidence
%  Similar to ReCo, their results on BoolQ classify YES/NO answers, and do not analyze the evidence. %(They also expand the passages of BoolQ but seem to indicate that its with random sentences?)

%https://arxiv.org/pdf/2006.12146.pdf\\
Besides TyDi, other MRC datasets have included a mix of boolean and non-boolean questions.  The Natural questions dataset has only around 1\% boolean questions.\cite{naturalQuestions}
MS MARCO \cite{bajaj2018ms} is a reading comprehension dataset that was created from the Bing search engine query logs with a substantial number of the boolean questions, but it does not provide evidence spans for explanation.

The passages from BoolQ were annotated by \cite{thorne2021evidencebased} to obtain supporting evidence for use in a claim verification task similar to FEVER \cite{Thorne18Fever,Thorne18Fact}. It is unclear whether this evidence is useful for question answering because they transform the questions into claims for the claim verification task. Boolean questions have also been used in other tasks such as conversational question answering \cite{Choi2018QuACQA,Reddy2019CoQAAC}. 
%QuAC \cite{Choi2018QuACQA} is and CoQA \cite{Reddy2019CoQAAC}  are QA dataset designed for dialog.  Given a passage and a conversation history, the task is to answer the next question in the conversation. The dialog may also contain boolean questions that require extracting the relevant evidence  form the input  passage. Thus our work can contribute to advancing SOTA in this area. 

%Evidence in other tasks e.g. FEVER claim verification.?
\section{Data}
\label{sec:data}

\begin{table}[t]
\small
\centering
    \begin{tabular}{c|c|c|c|c}
    \toprule
    \textbf{Dataset} & YES & NO & NA & Total \\
     \hline
     \midrule
     \tydi Train & 404 & 98 & 60 & 562 \\
     \tydi Dev & 52 & 25 & 1 & 78 \\
     BoolQ Dev & 1300 & 763 & 315 & 2378 \\
     \bottomrule
    \end{tabular}
    \caption{Number of boolean instances in the TyDi English and BoolQ datasets annotated for evidence. A breakdown of Yes, No, and Unanswerable (NA) questions is provided for each dataset.}
    \label{tab:data-stats}
\end{table}

We use two datasets that contain boolean questions in our analysis and experiments: \tydi \cite{clark-etal-2020-tydi} and BoolQ \cite{clark-etal-2019-boolq}. We chose these datasets due to their proportion of boolean questions and non-restrictive licences. In the following subsections we describe how we use and expand the datasets. For a detailed description of how the boolean questions were created we direct interested readers to \cite{naturalQuestions, clark-etal-2019-boolq, clark-etal-2020-tydi}. The  sizes of the portions of the datasets that we use are shown in Table~\ref{tab:data-stats}. Both datasets provide a train and development set that are available to download. The test sets are kept blind for leaderboard submissions, thus our results are reported on the boolean portion of the English dev sets.

\subsection{T{\small Y}D{\small I} QA}

\tydi is a multilingual MRC dataset containing questions in multiple languages. 10\% of the questions are boolean. The \tydi boolean question annotations include the paragraph the answer is found in and a label: YES or NO. A boolean question can also be unanswerable.
Each question in the dataset is paired with a supporting Wikipedia document, including the offsets of all its passages.
The \tydi dataset provides a paragraph-length answer, called the \textit{passage answer}, that supports the answer to both boolean and factoid questions. 
The dataset also provides a \textit{short answer} for non-boolean questions. %(found within the passage answer), which is the text span that answers non-boolean questions and which is absent from boolean questions.
It is important to note that this dataset is extremely skewed: 80\% of answerable boolean questions are YES.
In this paper we explore the English boolean questions which have an answer according to the \tydi annotations.

We provide our annotators with the passage answer and the preceding and following paragraphs from the Wikipedia document.
%We provide our annotators with the Wikipedia paragraph marked as containing the answer by TyDi and the surrounding paragraphs before and after it \avi{I don't understand this sentence.}.
We annotated the full \tydi English dataset for evidence consisting of 640 instances in total.

\subsection{BoolQ-X}

BoolQ is a dataset consisting of over 18k questions that are all boolean and answerable. BoolQ provides the question, a boolean answer as YES or NO, and a paragraph-length passage (around 100 words on average.)
In contrast to \tydi which is a document level MRC task, BoolQ is only a passage level MRC task.
In order to make the BoolQ data more useful in conjunction with the \tydi data, we locate (by title) the source Wikipedia article for each BoolQ passage in the 2018-12-20 Wikipedia dump.
(Note that the original BoolQ passages had been retrieved from the "live" Wikipedia.)
This mismatch in Wikipedia versions caused a loss of some questions in the original dataset from 3.2k to 2.4k dev instances.

% In this paper, we explore the BoolQ dataset in a zero-shot setting and thus only annotate the development set.

%Therefore, we convert the BoolQ data to the document level using the TyDi json format.
In order to obtain more useful context for annotation, we split each article into 100 word chunks, and provide the annotators with a 500 word context passage centered on the original BoolQ passage. We refer to our modification of the BoolQ dataset as BoolQ-X in order to clearly distinguish that our results are on BoolQ-X and not the original BoolQ dataset.
\section{Evidence Annotation}
\label{sec:annotation}

\begin{table*}[t]
    \centering
     \small
    \begin{tabular}{l|l|l}
    \toprule
	&	\tydi	&	BoolQ-X	\\
	\midrule
Majority of annotators agree with Original Label &	98 (77.2\%)	&	210 (85.0\%)	\\
One or more annotators had conflicting YES and NO sentences by the same annotator &	24 (18.9\%)	&	16 (10.5\%)	\\
Two or more annotators had conflicting YES and NO sentences by the same annotator &	14 (11.0\%)	&	12 (4.9\%)	\\
Conflicting YES and NO sentences by two or more different annotators & 34 (26.8\%) & 40 (16.2\%) \\
Question was marked unanswerable by majority of annotators &	17 (13.4\%)	&	48 (19.4\%)	\\
Majority of annotators flipped YES <-> NO compared to original \tydi annotation	&	7 (5.5\%)	&	16 (6.5\%)	\\
\midrule
Total Annotations	&	127	&	247	\\
\bottomrule
 \end{tabular}
    \caption{Analysis of \tydi and BoolQ-X questions that were annotated by all annotators.}
    \label{tab:annotation_analysis}
\end{table*}

We used an in-house annotation tool to annotate the evidence that explains the YES/NO answer to the \tydi and BoolQ-X questions. The annotator is provided with the question and passages. The passage displayed to the annotator were divided into sentences. (See Figure~\ref{fig:annotation_tool} in the Appendix for a snapshot of the annotation tool.) Although the original datasets provide us with a YES/NO answer, we do not show the label to the annotators and ask them to provide us with this information themselves. Given the question and passage, our annotation task was as follows:

\begin{enumerate}
\item Is there an answer? If not, it is "Unanswerable" (See Section~\ref{sec:unanswerable} for further discussion).
\item If there is an answer, highlight 1-3 spans for YES and NO. Label each span: YES, NO, WEAK\_YES, or WEAK\_NO. A span is a full sentence or a clause(s) in the sentence.
\item If the answer is WEAK, provide a reason:
\begin{itemize}[leftmargin=0cm]
    \item EXTRA\_INFO: The span contains extra words that don't contribute to the evidence.
    \item MISSING\_COREF: The first mention appears in the title or earlier in the passage.
    \item PARTIAL\_ANSWER: The evidence does not satisfy the question completely.
    \item QUESTION\_AMBIGUITY: Ambiguity in the question such as poor vocabulary, misspellings, and subjective meaning of a word
    \item ANSWER\_AMBIGUITY: Ambiguity in the answer span such as poor vocabulary, misspellings, and subjective meaning of a word
    \item OTHER: Any explanation not yet covered.
\end{itemize}
\end{enumerate}

The data was annotated by 10 full time annotators. We had two pilot rounds to ensure understanding of the task.  All 10 annotators annotated the \tydi dataset and 5 of the annotators annotated the BoolQ-X dataset. We assigned an overlap of 10\% of the \tydi training, the full \tydi dev set, and 10\% of the BoolQ-X dev set to be annotated by all the annotators to compute inter-annotator agreement.

% We compute agreement for the following: 1) Sentence level YES/NO evidence agreement among annotators, 2) Document level YES/NO agreement among annotators and 3) 

We compute inter-annotator agreement for the evidence on the sentence level using F1. We consider the annotators to agree if highlighted spans are in the same sentence  and have the same answer label. For the purpose of computing YES/NO sentence level agreement we ignored the WEAK labels by merging WEAK\_YES and YES and WEAK\_NO and NO. Finally, we compute agreement at the document level using accuracy. We consider the annotators to agree if they labeled one or more of the sentences in the document as evidence with the same label. %This is a useful measure because we instructed the annotators to pick 1-3 sentences to highlight and it is possible that more than three sentences exist. In other words, we want to ensure that the annotators highlight sentences that are valid evidence, but it is not necessary for all valid sentences to be highlighted. 

\textbf{Document level agreement to the original dataset:} The majority of annotators agreed with the original \tydi and BoolQ YES/NO labels indicating high agreement as shown in Table~\ref{tab:annotation_analysis}.

% We initially ran a small pilot of 30 questions where the authors reviewed all disagreements and came to a resolution. All the annotators were compared to the gold pilot annotations from the authors and on average the agreement was 53.4\% F1 at the sentence level and 66.6\% at the document level with the gold pilot annotations indicating moderate agreement. 
\textbf{Sentence and Document level inter-annotator agreement:} We compute agreement between each annotator and the annotator who was found to have the best agreement with the consensus in a pilot study. At the sentence level, the agreement was 55.3\% F1 in \tydi and 49.8\% in BoolQ-X. At the document level, the annotators agree on average 68.9\% in \tydi and 65.8\% in BoolQ-X for the full task. This indicates that there is moderate agreement for the full dataset but the annotators do vary on which sentences they highlight.

While the majority of annotators agreed with the original \tydi and BoolQ labels, we did find some inconsistencies. Table~\ref{tab:annotation_analysis} shows the occurrence of unanswerable questions, conflicting YES and NO labeled sentences for a single question by a single annotator, conflicting YES and NO sentences by different annotators, and the majority answer flipped from YES to NO or vice versa. We discuss these inconsistencies further in Section~\ref{sec:unanswerable} and ~\ref{sec:yn}.

\subsection{Data Release and Stats}

We prepare our annotated \tydi and BoolQ-X evidence spans for evaluation in a document level\footnote{BoolQ is a passage level task. See Appendix for how we prepared the BoolQ-X dataset for document level MRC.}  MRC system. During training we keep one evidence span for each question. During evaluation, we keep up to three evidence spans for each question. While we allowed the annotators to choose a portion of a sentence as the span, we only consider full sentences as evidence spans to avoid spans with overlapping offsets across annotators. If there are more than three sentences, (this can occur on occasion for the questions annotated by all annotators), we take the top three based on frequency and give preference to strong answers over weakly relevant answers. In the case of a tie we take the sentences earlier in the passage first. In the YES/NO experiments we consider the original \tydi and BoolQ labels to be correct even though on occasion our annotators disagreed with them. We do not include questions that our annotators considered unanswerable. Statistics about the evidence spans and YES/NO/Unanswerable labels can be found in Table~\ref{tab:span_stats} and Table~\ref{tab:data-stats}. Our data release %\footnote{upon acceptance} 
will include the data in \tydi format for easy replication of experiments.

\begin{table}[t]
    \centering
     \small
    \begin{tabular}{l|l|l}
    \toprule
Avg \# Evidence	&	\tydi	&	BoolQ-X	\\
	\midrule
sentences per document All & 2.3 & 1.5 \\
sentences per document Overlap & 3.6 & 2.9 \\
sentence length & 170 & 234 \\
span length & 112 & 132 \\
\bottomrule
 \end{tabular}
    \caption{Data Statistics for evidence annotations. Overlap refers to the data annotated by all annotators.}
    \label{tab:span_stats}
\end{table}
\section{Experiment Setup}
\label{sec:experiments}

In this section we describe our models and we evaluate the performance of evidence span extraction and YES/NO answer prediction.  We also demonstrate that incorporating supporting evidence in the MRC training improves %significantly 
the evidence span extraction performance. 

Our approach is based on the  XLM-Roberta language model \cite{conneau2020unsupervised}, as implemented in \cite{wolf-etal-2020-transformers}. We use separate models for evidence span extraction and for answer classification (YES/NO). The evidence span extraction is similar to the short answer extraction of \cite{alberti2019bert}. 

In our experiments we used the \tydi training set to train all systems and we evaluated on the annotated evidence spans in three experiments: \tydi dev, \tydi cross-validation (CV),  and BoolQ-X. The \tydi dev set only contains 77 answerable English boolean questions. To compensate for the small size of the \tydi dev dataset, we also perform 6-fold cross-validation using the 562 training examples and we average the result over all 6 folds. %The evaluation size for each fold is similar to the size of the actual dev set. 
The BoolQ-X experiments are zero-shot: the models are trained on the \tydi train set and evaluated on all the 2,063 answerable BoolQ-X annotated evidence spans. 
% Since our focus is on the quality of the extracted evidence spans, our experiments are performed on the examples where the annotators indicated there is an answer. 

\subsection{Evidence Span Extraction}

% In this work we show that a simple YES/NO answer is not sufficient and evidence is necessary to address the information needs in boolean questions. Text classification models that were used in prior work to provide a YES/NO answer based on a short supporting passage are no longer suitable ~\cite{clark-etal-2019-boolq}. 

% We address the boolean questions using a reading comprehension model to extract the evidence span from a large supporting document. Our reading comprehension systems uses the XLM-R Large language model \cite{conneau2020unsupervised} in a setup similar to \cite{alberti2019bert}.

%In this section we show that incorporating evidence into MRC models is necessary to address the information needs for Boolean questions. Our reading comprehension systems uses the XLM-R Large language model \cite{conneau2020unsupervised} in a setup similar to \cite{alberti2019bert}. In addition to obtaining evidence spans, we apply an evidence span classifier to produce a YES/NO answer to the question. The evidence span classifier is an XLM-R Large classification model.

We train an MRC system with all the examples in the \tydi training set, including non-boolean questions. During training, the MRC objective is to extract a span of text. This span will be the evidence when the question is boolean and a short answer otherwise. 
We investigate different strategies for incorporating evidence for boolean questions during MRC training.
%and we measure the F1 score of the extracted evidence on the Boolean questions subset in Table~\ref{tab:evidence-results}.  

\paragraph{Baseline (BASE):} A standard MRC system using out-of-the-box training data consisting of both boolean and short answer questions.
 None of the boolean questions have evidence spans during training and the YES/NO labels are ignored as in prior work~\cite{Zhang2021PoolingformerLD,wang2020answer,Wang2020ClusterFormerCS}.

\paragraph{Boolean Passage Evidence Span (BPES):} An MRC system baseline where gold \textit{passage} spans are used to approximate evidence spans for boolean questions during training. Both \tydi and BoolQ provide the passage where the answer is found and  we use these annotations to compensate for the lack of short answer evidence spans in the BASE system. 

\paragraph{Boolean Evidence Annotated Spans (BESA):} An MRC system where we use our sentence-level annotated evidence spans for the boolean questions during training. 
%We incorporate our evidence annotations into the TyDi QA train set.
%in place of the YES/NO answer labels following the same approach as our evaluation dataset described below. 
Since our evidence-span annotations are only on the English boolean training data, the BESA model is initialized from the BPES model during fine-tuning.
\\
\\\noindent
\textbf{ Results:} The evidence span extraction results are shown in Table~\ref{tab:evidence-results} using the F1 score computed with the official \tydi evaluation script. Other than \tydi dev, where the dataset is too small to find significant trends, the evidence span F1 increases with the quality of evidence span provided during training. The baseline BPES system, which uses a rough approximation of evidence spans with the gold passage, improves the performance by roughly 2 F1 points on both \tydi CV and BoolQ-X. The best performance is obtained using the BESA system (our evidence span annotations) with an increase of over 6 F1 points on both \tydi CV and BoolQ-X.

% \begin{table}[t]
%     \centering
%     % \small
%     \begin{tabular}{c|c|c||c|c||c|c}
%     & \multicolumn{2}{c||}{TyDi CV} & \multicolumn{2}{c||}{TyDi Dev} & \multicolumn{2}{c}{BoolQ-X Dev} \\
% 	&	SA	&	PA	&	SA	&	PA	&	SA	&	PA	\\
% 	\hline
% BASE	&	29.3	&	45.9	&	34.6	&	53.6	&	25.8	&	49.6	\\
% BPES	&	31.2	&	47.9	&	32.8	&	51.9	&	27.8	&	51.1	\\
% BESA &	35.5	&	46.8	&	35.7	&	46.7	&	32.1	&	46.3	\\ \end{tabular}
%     \caption{Evidence Results}
%     \label{tab:evidence-results}
% \end{table}

\begin{table}[t]
    \centering
    \small
    \begin{tabular}{c|c||c||c}
    \toprule
    & \multicolumn{1}{c||}{\tydi} & \multicolumn{1}{c||}{\tydi} & \multicolumn{1}{c}{BoolQ-X} \\
    & CV & Dev & Dev \\
    \midrule
BASE	&	29.3	&	34.6	&	25.7	\\
BPES	&	31.2	&	32.8	&	27.6	\\
BESA &	35.5		&	35.7	&	31.9	\\ 
\bottomrule
\end{tabular}
    \caption{Evidence Span Extraction Results (F1) for \tydi cross validation (CV) and \tydi and BoolQ development sets}
    \label{tab:evidence-results}
\end{table}

\subsection{Answer Classifier Experiments}

After the evidence span is extracted, we use the answer classifier to produce a YES/NO answer for the boolean questions at the document level.
We use the gold passages for all boolean questions in the \tydi dataset to train the answer classifier and we show the YES/NO F1 score in Table~\ref{tab:yn-results}.
The answer classifier training data contains examples from all \tydi languages while the evaluation is on the English subset of \tydi, corresponding to our annotations. 
Each example in the evaluation subset has a single YES/NO answer. 
% In cases where our annotators have marked conflicting YES and NO answers, we kept the answer that was consistent with the TyDi original dataset.

%We use the TyDi QA YES/NO annotation as gold standard where each boolean question has a single answer.

%Training with full passage provides additional context to the system such as  co-reference resolution.
%which a user does not want to have highlighted for readability. 
 We compare the results with the majority baseline of always answering YES. This is a strong baseline because both datasets are skewed (see Table \ref{tab:data-stats}.) %for the TyDi QA dataset (where there is an extreme skew towards YES) but also for BoolQ-X, since the majority of boolean questions have the answer YES.

% We apply the evidence span classifier on the extracted evidence span only and on the passage containing the extracted evidence. In our experiments we see that the evidence span classifier is more effective when it is given the entire passage around the extracted evidence span, indicating that a larger context is beneficial.
% This result is consistent with prior work on the original BoolQ dataset ~\cite{clark-etal-2019-boolq} where the input is also a passage.
% We expect the full paragraph provides additional context to the system such as important co-reference resolution which a user does not want to have highlighted for readability. 
We tested the answer classifier using the passage containing the extracted evidence span (BESA) and also the gold passage annotations from both \tydi and BoolQ-X. %(Passage boundaries are provided with both TyDi QA and BoolQ-X datasets) 
Using the gold passage as context for the answer classifier assumes the extracted evidence spans are always correct, and this is an upper bound for our results. 

%We ran experiments using the gold passage annotations instead of the passage of the extracted evidence span to obtain an upper bound for the evidence span classifier. This assumes that the system predicted passages are always correct. 

\textbf{Results:} We notice that the evidence span classifier is close to the upper bound on \tydi dataset, and there is little room for improvement, likely due to the skewed distribution of YES/NO answers.   On the BoolQ-X dataset there is a larger difference (8 F1 points) when applying the classifier on system passages compared to gold passages. Our system passage results outperform the strong majority baseline for both datasets. This indicates that better performance of evidence span extraction will also improve the YES/NO answer prediction.

We also experimented with using only the predicted evidence span as input to the answer classifier and obtained weaker results. We conjecture that the full paragraph containing the evidence span provides additional context to the system such as co-reference resolution. 

\begin{table}[t]
    \centering
    \small
    \begin{tabular}{c|c||c||c}
    \toprule
    & \tydi  & \tydi & BoolQ-X \\
    & CV & Dev & Dev \\
	\midrule
Majority & 80.8	&	77.1	&	63.0			\\ 
Gold Passage (UB)	&	85.5	&	85.7	&	78.9			\\
\midrule
Sys Passage	&	82.0	&	87.0	&	74.3			\\
Evidence Span &	81.2	&	85.0	&	70.1			\\
\bottomrule
\end{tabular}
    \caption{ YES/NO Results (F1) for \tydi cross validation (CV) and \tydi and BoolQ development sets. The evidence span classifier is applied on: 1. System Passage from BESA system containing the evidence span; 2. Gold Passage. 3. Extracted evidence span from BESA system; }
    \label{tab:yn-results}
\end{table}

\section{Boolean Evidence Spans - User Study}
\label{sec:userstudy}

Our experiments in the previous sections evaluated the performance of our systems against the gold answers and showed that the BESA model, which includes evidence spans at training time, performs best. However, %our results are still modest and 
it is difficult to interpret
%how the results relate to
how useful an end user would find the evidence. In this section, we explore whether users find the highlighted evidence to be informative for each of the models via a user study. This helps determine which model provides the most informative explanation as evidence to answer the question. 

We perform a user study on the full answerable \tydi dev set of 77 questions and an equal portion of the BoolQ-X dev set. An example of the user study interface is provided in the Appendix Figure~\ref{fig:user_study_tool}. Our findings from the user study are shown in Table~\ref{tab:userstudy}.

\begin{table*}[t]
    \centering
     \small
    \begin{tabular}{c|c|c|c|c|c|c} 
    \toprule
	&	\multicolumn{3}{c|}{\tydi} &	\multicolumn{3}{c}{BoolQ} \\
	& BASE & BPES & BESA & BASE & BPES & BESA \\
	\midrule
Relevant	&	45\%	&	42\%	&	55\%	&	32\%	&	42\%	&	58\%	\\
Irrelevant	&	29\%	&	37\%	&	27\%	&	39\%	&	27\%	&	16\%	\\
Weakly Relevant	&	26\%	&	21\%	&	17\%	&	29\%	&	31\%	&	26\%	\\
\midrule
Relevant/Weak and in  Gold	&	38\%	&	36\%	&	37\%	&	26\%	&	34\%	&	38\%	\\
Relevant/Weak but \textit{not in} Gold	&	33\%	&	27\%	&	35\%	&	35\%	&	39\%	&	46\%	\\
\midrule
Mean Reciprocal Rank	&	0.79	&	0.77	&	0.88	&	0.71	&	0.77	&	0.87	\\
\midrule
Poor Highlighting	&	8\%	&	14\%	&	3\%	&	32\%	&	20\%	&	5\%	\\
% \midrule
% Total &	\multicolumn{3}{c|}{\textbf{77}} &	\multicolumn{3}{c}{\textbf{77}} \\
\bottomrule
 \end{tabular}
    \caption{User Study analysis of the evidence spans produced by each system averaged across the three participants for the 77 \tydi and BoolQ-X questions from the development sets.}
    \label{tab:userstudy}
\end{table*}

% OLD W/ ERRORS
% \begin{table*}[t]
%     \centering
%      \small
%     \begin{tabular}{c|c|c|c|c|c|c} 
%     \toprule
% 	&	\multicolumn{3}{c|}{\textbf{TyDi}} &	\multicolumn{3}{c}{\textbf{BoolQ}} \\
% 	& BASE & BPES & BESA & BASE & BPES & BESA \\
% 	\midrule
% Relevant	&	45\%	&	42\%	&	55\% &	45\%	&	40\%	&	46\%	\\
% Irrelevant	&	29\%	&	37\%	&	27\%	&	28\%	&	27\%	&	29\%	\\
% Weakly Relevant	&	26\%	&	21\%	&	17\%	&	28\%	&	33\%	&	25\%	\\
% \midrule
% Relevant/Weak and in gold	&	38\%	&	36\%	&	37\%	&	23\%	&	25\%	&	26\%	\\
% Relevant/Weak but \textit{not in} gold	&	33\%	&	27\%	&	35\%	&	49\%	&	48\%	&	45\%	\\
% 	\midrule
% Mean Reciprocal Rank	&	0.79	&	0.77	&	0.88	&	0.77	&	0.75	&	0.82	\\
% \midrule
% Poor Highlighting	&	8\%	&	14\%	&	3\%	&	20\%	&	20\%	&	19\%	\\
% % \midrule
% % Total &	\multicolumn{3}{c|}{\textbf{77}} &	\multicolumn{3}{c}{\textbf{77}} \\
% \bottomrule
%  \end{tabular}
%     \caption{User Study analysis averaged across the three participants for the 77 TyDi and BoolQ questions from the development sets.}
%     \label{tab:userstudy}
% \end{table*}

The user study was completed by three of our annotators, each analyzing all 77 questions from each dataset. We provided these users with an interface that contained the output from each of our three models: BASE, BPES, and BESA. The output was randomized and anonymized (the users only saw PRED A, B, C instead of the model name) ensuring that the users did not know which system the output originated from. The study explored two aspects of the models which the users had to answer: 1) For each model, is the highlighted evidence span relevant, weakly relevant, or irrelevant for answering the question? 2) Rank the passages based on preference of which highlighted span is best. In other words, which highlight would you (the user) want to see for the given question? Asking these two distinct questions regarding relevance and rank is important because even if all the answers are irrelevant, some evidence spans may be more appropriate than others based on their pertinence to the question.\footnote{The highlighting of inappropriate spans can lead to embarrassing answers.} While in most cases the top rank is correlated with relevance, in our study the rank is designed to compare the highlighted spans of the three models to each other. 

We first examine the relevance of the highlighted evidence spans and find that all users picked BESA as having the most relevant evidence spans and the least irrelevant evidnece spans as shown in Table \ref{tab:userstudy} for both datasets. Weakly relevant evidence spans were more common in BoolQ-X, which is zero-shot. Further, in both datasets there are many evidence spans that are relevant but were not found in the gold evidence span annotations. These are evidence spans found by the system (which saw the full document) while the annotators only received the candidate passage and its surrounding paragraphs. We find that for all the models the number of relevant evidence spans not in gold is comparable to or larger than the number of relevant evidence spans in gold for both datasets. This indicates that our models predict relevant evidence spans considerably more often than reflected by our evaluation shown in Table \ref{tab:evidence-results}.  

 Next we examine the ranking of the three models. The rank given by the users was from 1-3 where the best evidence span is given a rank of 1 and the worse span is given a rank of 3. Multiple evidence spans could have the same rank in cases of a tie. If there is a tie the count would continue by skipping the next position (e.g. 1 1 3 instead of 1 1 2). We discouraged ties when the highlights were not identical, although there are some cases where the users considered two different evidence spans to be equally informative and they were assigned the same rank. We used Mean Reciprocal Rank (MRR) to compare the ranking of the three models. In all cases the annotators preferred BESA over the other models. 
 %Both BASE and BPES do not generate good approximations for evidence based on user preference. 
The MRR scores reflect our results shown in Table~\ref{tab:evidence-results}: BPES is preferred for BoolQ-X and BASE is preferred for \tydi. The average MRR is shown in Table \ref{tab:userstudy}.

%  This further indicates that we actually perform better than the results in Table~\ref{tab:evidence-results} illustrate. At a minimum we have XX\% more correct for TyDi and XX\% more correct for BoolQ with the most conservative annotator. 

% \begin{table}[t]
%     \centering
%      \small
%      \setlength{\tabcolsep}{3.3pt}
%     \begin{tabular}{p{1.9cm}|p{5.5cm}} 
%     \toprule
% \textbf{Poor } & \textbf{Good} \\
% \textbf{Highlighting} & \textbf{Highlighting} \\

% \midrule
% The Amazing Spider-Man 2 & It is also the second and final film in "The Amazing Spider-Man" franchise. \\
% \midrule
% Starbucks & Since 2003, they have been a subsidiary of American coffeehouse chain Starbucks. \\
% \midrule
% Cargo is a 2017 & Cargo is a 2017 Australian post-apocalyptic drama thriller film directed by Ben Howling and Yolanda Ramke \\
% \bottomrule
%  \end{tabular}
%     \caption{Examples of poor and good highlighting generated by various models as evidence. The full examples can be found in Appendix \ref{sec:appendix-user} }
%     \label{tab:highlighting}
% \end{table}

 We also asked the annotators to specifically record any predictions that had poor highlighting span boundaries. Examples of poor highlighting span boundaries include spans in which only one or two words were highlighted, and cases where highlighting broke off mid-word or phrase. Some examples of good and poor highlighting are available in Appendix \ref{sec:appendix-user}.
%  shown in Table~\ref{tab:highlighting}. 
Table~\ref{tab:userstudy} shows that poor highlighting span boundaries was most infrequent for the BESA model for both datasets. In general, highlighting was poorer for BoolQ-X. This is mostly likely due to the zero-shot transfer from {\sc TyDi QA}. 
 
 We also provided the users with the question and answer paragraph with and without highlighting and asked them in general whether they prefer to have the paragraph with or without the highlighting. All the annotators said they preferred the highlighting. This indicates that adding highlighting to current systems that provide a YES/NO answer with a full paragraph would be helpful to the user.
 
%The responses were stored in an Excel spreadsheet.

% Analysis of user study goes here...

% To compute overall ranking we added the ranks for each system and compared the overall rankings per annotator. A lower overall score indicates that system most often had a better ranking. In all cases all the annotators likes our BESA model the best for both TyDi and BoolQ. For TyDi, both BASE and BPES were similar and in BoolQ BASE was better than BPES.

% Relevance etc ...

\section{Analysis}

\subsection{Unanswerable Questions}
\label{sec:unanswerable}

During the annotation task our annotators found some questions to be unanswerable even though all of our annotations were performed on question-answer pairs that had an answer of YES or NO in the original datasets. %In our full dataset, 52 (8.1\%) TyDi and 314 (12.7\%) BoolQ, 
9\% of \tydi and 13\% of BoolQ-X
questions were marked as unanswerable as shown in Table~\ref{tab:data-stats}. Table~\ref{tab:annotation_analysis} shows the number of questions that were marked unanswerable when all annotators saw the data. %Among the data that was annotated by all the annotators, 64/154 (41.2\%) of the TyDi and 26/247 (10.5\%) of the BoolQ documents were marked as unanswerable by two or more annotators. 

We found that questions tend to be unanswerable for two main reasons. First, sometimes the question is not answered in the page.  While this could indicate that the answer is NO, it is not clear. For example, consider the question ``\textit{Do US Nintendo 64 games work in Australia?}" from the BoolQ dataset. While the passage provided was about Nintendo 64 games, Australia was not mentioned within the supporting document, so this question cannot be answered. Second, sometimes the the inappropriate context was provided for the question. For example, consider the question \textit{``Is Major General Douglas dead?''} from the \tydi dataset. This question is provided with the incorrect document about ``Douglas Haig, 1st Earl Haig'' instead of the document about ``Major General Alexander Douglas Campbell''. 
%The rest of our experiments and analyses exclude the unanswerable questions.

\subsection{Conflicting Answers}
\label{sec:yn}

Prior work on boolean questions makes the assumption that the answer is either YES or NO. However, it is possible to have a question where there are \textit{conflicting answers} to a boolean question that are both correct. Consider the following example from the BoolQ dataset: \\

\begin{minipage}{.95\columnwidth}
\small
\noindent\textit{Question}: Do you need to say check in chess? \\ \\
\textit{Title}: Check (Chess) \\ \\
\textit{Passage}: A player must get out of check, if possible, by interposing a piece between the threatening piece, or block the check with another piece. A king cannot itself directly check the opposing king, since this would place the first king in check as well. A move of the king could expose the opposing king to a discovered check by another piece, however. \textbf{In informal games, it is customary to announce "check" when making a move that puts the opponent' s king in check}\textcolor{blue}{\textbf{\textsuperscript{YES}}}. \textbf{In formal competitions, however, check is rarely announced}\textcolor{darkgreen}{\textbf{\textsuperscript{NO}}}...
\end{minipage}
\\

\noindent The above question has both YES and NO answers in the passage, depending upon the setting (informal game vs. formal competition) in which the chess game is played. This example illustrates that conflicting answers often occur when a question is under-specified. The answer may depend on certain demographics, a time period, or other characteristics relevant to the question topic. Under-specified questions are very common. For example, a person may ask the question "Can I get the COVID vaccine?" which can be YES or NO depending on when they are asking, their age, occupation, location, or other health factors.  Another example of conflicting answers in a single passage is shown in Figure~\ref{fig:bool-examples}.2.

%We examine the occurrence of conflicting answers to a single question in the TyDi and BoolQ datasets even though the labelling process does not incorporate this scenario and found that it does exist in both datasets. 

The frequency of conflicting answers in \tydi and BoolQ-X from the same annotator and from multiple annotators is shown in Table~\ref{tab:annotation_analysis}. Conflicting answers occur in both datasets, with more frequency in \tydi. The need for more information to answer the question can be addressed using dialog with the user, an interesting area to explore in future work. 

% We examined the overlapping data in TyDi and BoolQ that was annotated by all annotators for YES and NO answers in the passage. There were multiple annotators that labeled sentence(s) as YES and multiple annotators that labeled sentence(s) as NO in both datasets; 24.4\% of TyDi and 18.6\% of BoolQ. Further, often \textit{a single annotator} labeled different sentences in the same passage as YES and NO as shown previously in Table~\ref{tab:annotation_analysis}. This indicates that both YES and NO answers to a single question occurs in both datasets, with more frequency in TyDi. 

\subsection{Weakly Relevant Answers}
\label{sec:weak}

\begin{table}[t]
    \centering
     \small
    \begin{tabular}{l|l|l}
    \toprule
	&	\tydi Weak	&	BoolQ-X Weak	\\
	\midrule
Partial Answer	&	113	&	193	\\
Missing Coref	&	118	&	309	\\
Answer Ambiguity	&	184	&	328	\\
Other	&	15	&	3	\\
Extra Info	&	28	&	50	\\
Question Ambiguity	&	36	&	39	\\
\# Weak Sentences	&	515 (37.6\%)	&	922 (27.7\%)	\\
\# Weak Documents	&	69	(23.9\%)	& 306	(12.4\%) \\
 \end{tabular}
    \caption{Analysis of reasons for weakly relevant sentences in the full \tydi and BoolQ-X annotations. A sentence is considered weak if one or more annotators marked it as weak. A document is considered weak if all the annotations in the document are weak.}
    \label{tab:weak_analysis}
\end{table}

In some cases there is evidence in the passage that is relevant to the answer but does not completely answer the question (See example in Figure~\ref{fig:bool-examples}.1). We consider such evidence to be \textit{weakly relevant}. During annotation, we allowed the annotators to indicate that the evidence they selected was weak by indicating that it was a WEAK\_YES or a WEAK\_NO. When a weak label was selected we also requested that they provide a reason that it was weak as described in Section~\ref{sec:annotation}\footnote{Note we did not strictly enforce the annotators to provide a reason for weak annotations, thus the number of weak sentences is more than the sum of the different reasons.}. Table~\ref{tab:weak_analysis} shows the occurrence of \textit{weakly relevant} sentence annotations and the reasons in the \tydi and BoolQ-X datasets. The count indicates that one or more annotators marked a weak annotation in a sentence, and the reason.
Different annotators could say a sentence had weak evidence but provide different reasons for their annotation. Weakly relevant evidence is more common in \tydi than BoolQ. The most common reason evidence is weak is due to answer ambiguity such as in the following example: \\

\begin{minipage}{.95\columnwidth}
\small
\noindent\textit{Question}: Can the Queen dissolve parliament? \\ \\
\textit{Title}: Dissolution of the Parliament of the United Kingdom \\ \\
\textit{Passage}: The last dissolution of Parliament was on 3 May 2017, to make way for the general election to be held on 8 June 2017. [7] It dissolved after a two-thirds majority vote by the House of Commons, as required by the Fixed-term Parliaments Act. \textbf{30 March 2015 was the first time Parliament had ever dissolved automatically, as opposed to being dissolved by Royal Proclamation}\textcolor{cyan}{\textbf{\textsuperscript{WEAK\_YES}}} ...
\end{minipage}
\\

\noindent In the above example, the answer is ambiguous because it can be inferred that the Queen dissolved the parliament by Royal Proclamation even though it does not say so explicitly. Examples for the other reasons evidence is annotated as weak are available in the Appendix.

% \begin{figure}[t]
% \includegraphics[width=\columnwidth]{figures/besa_weak_analysis.pdf}
% \caption{Scores plotted based on relevance produced by the BESA system. The x-axis is normalized across the three types of relevance.}
% \label{fig:weak_analysis}
% \end{figure}

Our MRC system provides a confidence score for the answer.
We expect that the confidence score could be used to indicate the degree of relevance of the evidence to the answer.
During our user study we also asked the users to mark the evidence as \textit{relevant, irrelevant, or weakly relevant} as shown in Table~\ref{tab:userstudy}. %The annotators found the evidence on average 52\% Relevant, 33\% Irrelevant, and 14\% Weakly Relevant to the answer for the BESA system. 
We mapped the relevance to the confidence scores and found that on average the score for relevant answers is highest indicating there is some signal in the confidence scores. %In our experiments we predict evidence and the answer label. 
A user interface could indicate the confidence by making the highlighted text lighter if it is \textit{weakly relevant} and darker if it is \textit{relevant}. We leave further exploration of using confidence to indicate degree of relevance as future work.
\section{Conclusion}

In this paper we explore the importance of providing a YES/NO response \textit{and} evidence for boolean questions to the end user of machine reading comprehension systems. We provide detailed evidence annotations on top of the boolean questions in the popular BoolQ and \tydi datasets. We analyze several aspects that have been previously overlooked: the occurrence of supporting evidence, the degree of relevance of answers, and existence of conflicting answers in a single passage. We will be releasing our annotations for research purposes. 
% \footnote{Upon acceptance}. 

We show that providing evidence not only increases the model performance but also improves the user experience as measured in our user study. Our user study shows that receiving concise evidence in addition to the YES/NO answer and paragraph increases the confidence of the user in the answer. In the future we plan to explore degree of relevance and conflicting answers on a larger scale.

\bibliography{acl_bool}
\bibliographystyle{acl_natbib}

\appendix

\section{Annotation}
\label{sec:appendix}

\begin{figure*}[t]
\includegraphics{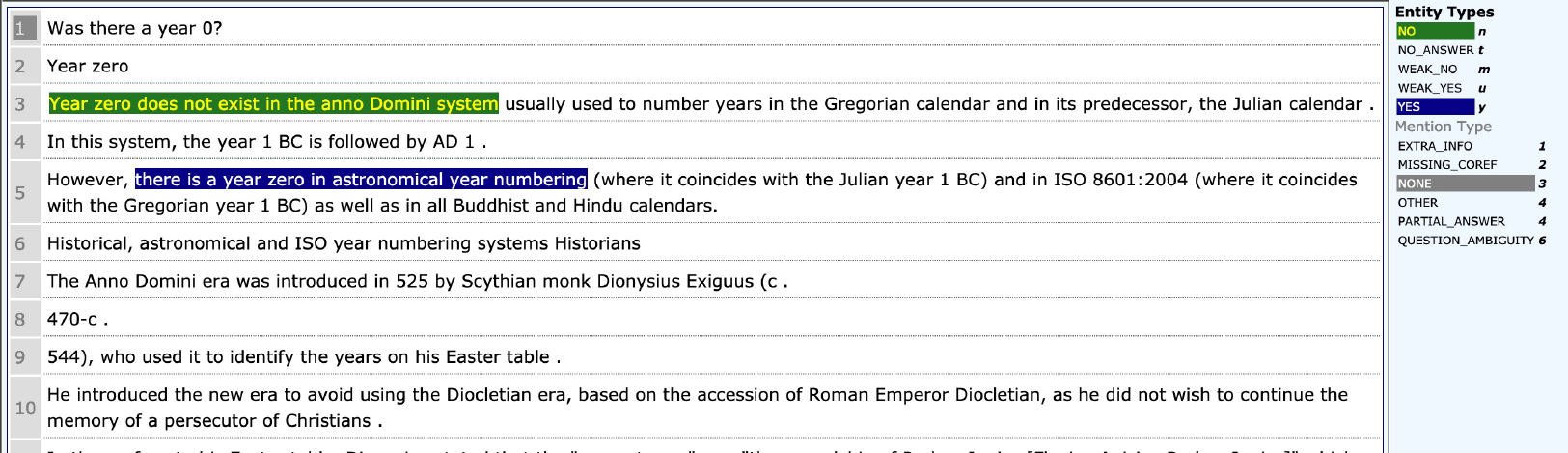}
\caption{Screenshot of the annotation tool used to find evidence.}
\label{fig:annotation_tool}
\end{figure*}

\subsection{Weakly Relevant Examples}

The following are examples of questions with Weakly Relevant evidence: Question Ambiguity, Missing Coref, Partial Answer, and Extra Info.

\subsubsection{Question Ambiguity}

\begin{minipage}{.95\columnwidth}
\small
\noindent\textit{Question}: Is Creole a pidgin of French?
\\ \\
\textit{Title}: French-based creole languages \\ \\
\textit{Passage}: A French creole, or French-based creole language, is a creole language (contact language with native speakers) for which French is the lexifier. Most often this lexifier is not modern French but rather a 17th-century koiné of French from Paris, the French Atlantic harbors, and the nascent French colonies. French-based creole languages are spoken natively by millions of people worldwide, primarily in the Americas and on archipelagos throughout the Indian Ocean. This article also contains information on French pidgin languages, contact languages that lack native speakers. ...
\end{minipage}
\\

\noindent The above \tydi question provides an example of \textit{question ambiguity} because a Creole is a pidgin with native speakers and does not refer to a specific language.

\subsubsection{Missing Coref}

\begin{minipage}{.95\columnwidth}
\small
\noindent\textit{Question}: Did the equal pay act pass? \\ \\
\textit{Title}: The Equal Pay Act of 1963 \\ \\
\textit{Passage}: The Equal Pay Act of 1963 is a United States labor law amending the Fair Labor Standards Act, aimed at abolishing wage disparity based on sex (see Gender pay gap). \textbf{It was signed into law on June 10, 1963}, by John F. Kennedy as part of his New Frontier Program. [1] ...
\end{minipage}
\\

\noindent The highlighted evidence in the above BoolQ example is \textbf{missing coref}. The Equal Pay Act is referenced in the first sentence.

\subsubsection{Partial Answer}

\begin{minipage}{.95\columnwidth}
\small
\noindent\textit{Question}: Are they actually singing in Grey's Anatomy? \\ \\
\textit{Title}: Song Beneath the Song \\ \\
\textit{Passage}: \textbf{A vocal coach was enlisted to help the cast.} [10] Music director Chris Horvath was recruited to arrange the selected songs for the cast. The arrangements took around two months, with vocals recorded over four days in February 2011.[7][12] Horvath praised the cast's response to the episode, noting that only four performers had "serious vocal talent," while some had "barely sung in the shower" before.[7] \textbf{Those with professional singing experience include Ramirez, who won a Tony Award for their role in the musical Spamalot, and Wilson, who appeared in the Broadway production of Caroline} ...
\end{minipage}
\\

\noindent The highlighted evidence spans in the above BoolQ example are \textit{partial answers} because the evidence does not satisfy the question completely. The evidence indicates that the actors did sing but does not say so explicitly. 

\subsubsection{Extra Info}

\begin{minipage}{.95\columnwidth}
\small
\noindent\textit{Question}: Does the UK have universal health care? \\ \\
\textit{Title}: List of countries with universal health care \\ \\
\textit{Passage}: 
Virtually all of Europe has either publicly sponsored and regulated universal health care or publicly provided universal healthcare. The public plans in some countries provide basic or "sick" coverage only, with their citizens being able to purchase supplemental insurance for additional coverage . \textbf{Countries with universal health care include Austria, Belarus,[55] Croatia, Czech Republic, Denmark, Finland, France, Germany, Greece, Iceland, Ireland, Italy, Luxembourg, Malta, Moldova,[56] the Netherlands, Norway, Portugal,[57] Romania, Russia, Serbia, Spain, Sweden, Switzerland, Turkey, Ukraine,[58] and the United Kingdom.[59]} ...
\end{minipage}
\\

\noindent In the above \tydi example countries other than the UK are included to ensure a complete evidence span is highlighted leading to \textit{extra information}. 

\subsection{Data Release}

We convert the BoolQ dataset, which is passage level, to a document level task for our experiments. 
We segment the original Wikipedia document into rough paragraphs by splitting it into sentences using the NLTK tokenizer. We keep roughly 500 characters in each paragraph, while ensuring a cutoff does not occur mid-sentence. We also ensure that the candidate answers are in a single paragraph. The document level version of BoolQ-X is available in \tydi format for easy replication of experiments.

\section{User Study}
\label{sec:appendix-user}

\begin{figure*}[t]
\includegraphics[width=\textwidth]{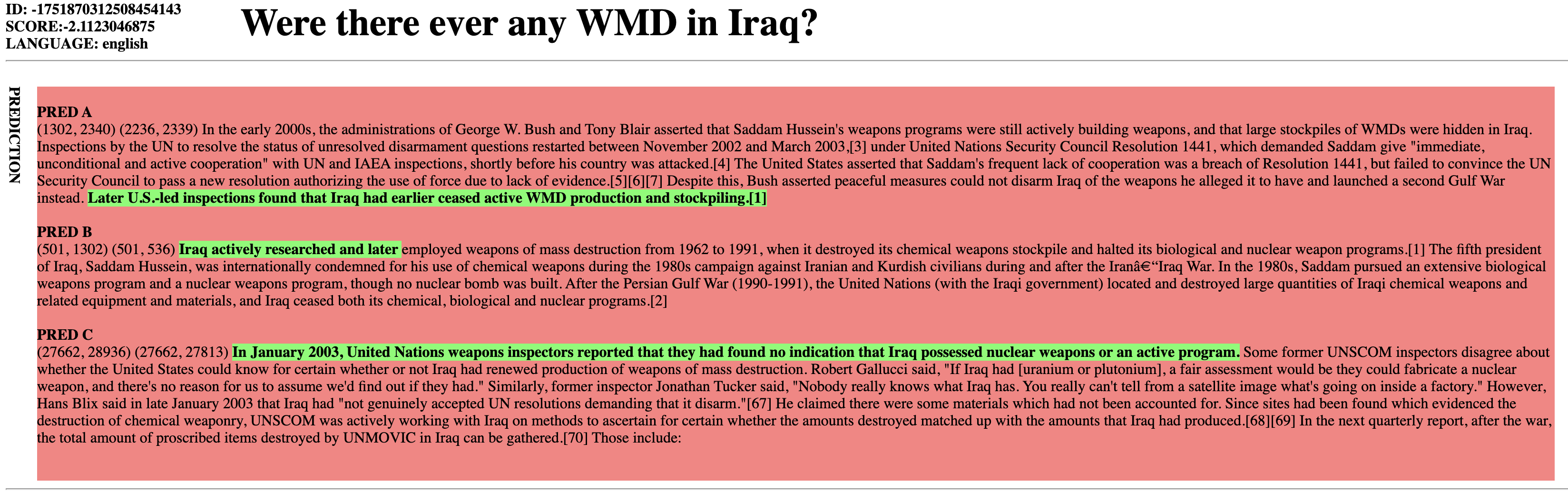}
\caption{Screenshot of the user study tool used to find evidence. In addition to the three randomized predictions shown here the users were also provided with the gold answers and full document text to optionally use if additional context was required.}
\label{fig:user_study_tool}
\end{figure*}

Figure~\ref{fig:user_study_tool} shows a screenshot of the tool the participants in the user study used to compare our various models. 

The following are examples of good and poor highlighting from the user study from the different models. The good highlighting is in \textbf{bold} and the poor highlighting is \underline{underlined}. (In some cases they overlap.)
\\ \\
\noindent
\begin{minipage}{.95\columnwidth}
\small
\textit{Question}: Is there a sequel to The Amazing Spider Man?
\\
\textit{Title}: The Amazing Spider-Man 2 \\
\textit{Passage}: \underline{The Amazing Spider-Man 2} (also released as The Amazing Spider-Man 2: Rise of Electro in some markets) is a 2014 American superhero film featuring the Marvel Comics character Spider-Man. The film was directed by Marc Webb and was produced by Avi Arad and Matt Tolmach. It is the fifth theatrical "Spider-Man" film produced by Columbia Pictures and Marvel Entertainment, and is the sequel to 2012's "The Amazing Spider-Man". \textbf{It is also the second and final film in "The Amazing Spider-Man" franchise}. The studio hired James Vanderbilt to write the screenplay and Alex Kurtzman and Roberto Orci to rewrite it.
\end{minipage}
\\ \\

\noindent
\begin{minipage}{.95\columnwidth}
\small
\noindent\textit{Question}: Is Starbucks and Seattle's Best same company?
\\
\textit{Title}: Seattle's Best Coffee \\
\textit{Passage}: Seattle's Best Coffee LLC is an American coffee retailer and wholesaler based in Seattle, Washington. \textbf{Since 2003, they have been a subsidiary of American coffeehouse chain \underline{Starbucks}}. Seattle's Best Coffee has retail stores and grocery sub-stores in 20 states and provinces and the District of Columbia. Sub-stores can also be found at many other businesses and college campuses, including JCPenney and Subway restaurants.
\end{minipage}
\\ \\

\noindent
\begin{minipage}{.95\columnwidth}
\small
\noindent\textit{Question}: Is Cargo an Australian movie? 
\\
\textit{Title}: Cargo (2017 film) \\
\textit{Passage}: \textbf{\underline{Cargo is a 2017} Australian post-apocalyptic drama thriller film directed by Ben Howling and Yolanda Ramke with a screenplay by Ramke} based on their 2013 short film of the same name.[1][2] The film stars Martin Freeman, Anthony Hayes, Susie Porter, and Caren Pistorius.[3]. It premiered at the Adelaide Film Festival on 6 October 2017 and was released in cinemas in Australia on 17 May 2018, worldwide except for Australia on 18 May 2018 by Netflix and on Netflix in Australia on 16 November 2018.[4]
\end{minipage}

\end{document}